\documentclass[pdflatex,sn-nature]{sn-jnl}

\usepackage{graphicx}%
\usepackage{multirow}%
\usepackage{amsmath,amssymb,amsfonts}%
\usepackage{amsthm}%
\usepackage{mathrsfs}%
\usepackage[title]{appendix}%
\usepackage{xcolor}%
\usepackage{textcomp}%
\usepackage{manyfoot}%
\usepackage{booktabs}%
\usepackage{algorithm}%
\usepackage{algorithmicx}%
\usepackage{algpseudocode}%
\usepackage{listings}%

\usepackage{multirow}
\usepackage{subcaption}
\usepackage{makecell}
\usepackage{float}
\usepackage{hyperref}

\theoremstyle{thmstyleone}%
\theoremstyle{thmstyletwo}%

\theoremstyle{thmstylethree}%

\begin{document}

\title[Article Title]{Achieving Deep Continual Learning via Evolution}


\author[1]{\fnm{Aojun} \sur{Lu}}\email{aojunlu@stu.scu.edu.cn}

\author[1]{\fnm{Junchao} \sur{Ke}}\email{kjc@stu.scu.edu.cn}

\author[1]{\fnm{Chunhui} \sur{Ding}}\email{chunhuiding9@gmail.com}

\author[1]{\fnm{Jiahao} \sur{Fan}}\email{fanjh@scu.edu.cn}

\author[1]{\fnm{Jiancheng} \sur{Lv}}\email{lvjiancheng@scu.edu.cn}

\author*[1]{\fnm{Yanan} \sur{Sun}}\email{ysun@scu.edu.cn}

\affil*[1]{\orgdiv{College of Computer Science}, \orgname{Science, Sichuan University}, \orgaddress{\street{1st Ring Road}, \city{Chengdu}, \postcode{610065}, \state{Sichuan}, \country{China}}}


\abstract{
Deep neural networks, despite their remarkable success, remain fundamentally limited in their ability to perform Continual Learning (CL). While most current methods aim to enhance the capabilities of a single model, Inspired by the collective learning mechanisms of human populations, we introduce Evolving Continual Learning (ECL), a framework that maintains and evolves a diverse population of neural network models. ECL continually searches for an optimal architecture for each introduced incremental task. This tailored model is trained on the corresponding task and archived as a specialized expert, contributing to a growing collection of skills. This approach inherently resolves the core CL challenges: stability is achieved through the isolation of expert models, while plasticity is greatly enhanced by evolving unique, task-specific architectures. Experimental results demonstrate that ECL significantly outperforms state-of-the-art individual-level CL methods. By shifting the focus from individual adaptation to collective evolution, ECL presents a novel path toward AI systems capable of CL.
}




\maketitle

\section{Introduction}\label{sec1}

\begin{figure*}[!t]
    \centering
    \includegraphics[width=1\linewidth]{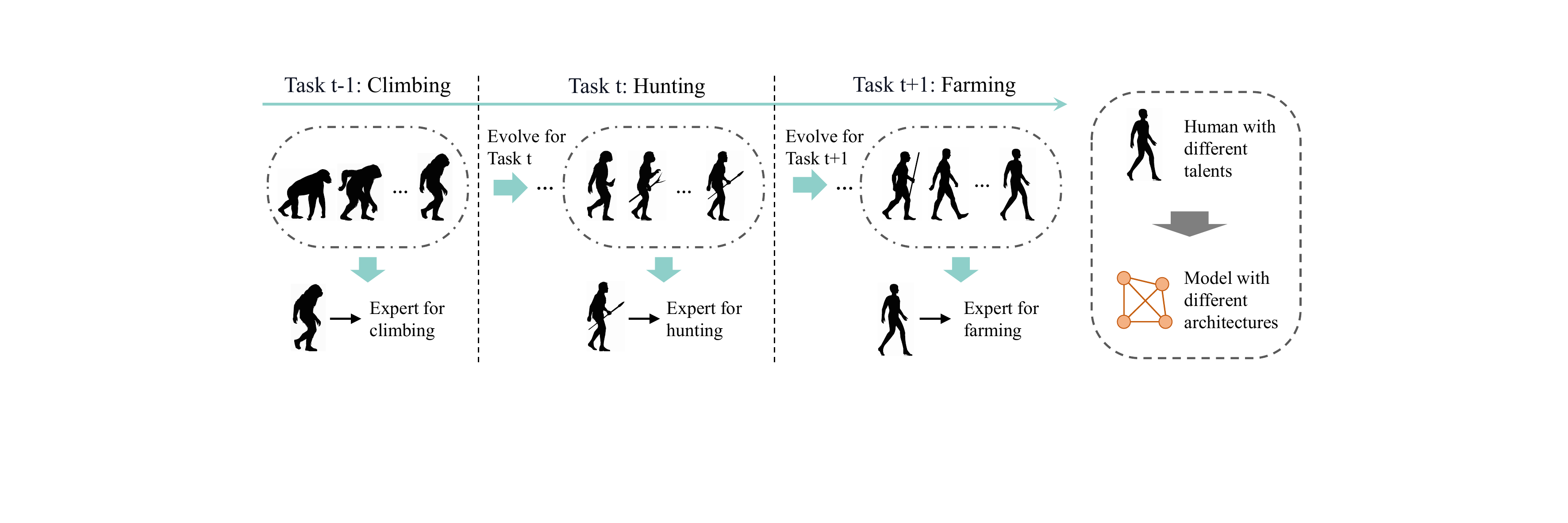}
    \caption{Inspired by the collective intelligence accumulation mechanism in humans, ECL proposes to achieve CL through an evolving population of models rather than a single one. This approach enables diverse model architectures to dynamically evolve via NAS, each adapting to specific tasks, while selecting the most promising architecture to acquire knowledge from the current task.}
    \label{fig:fig1}
\end{figure*}

Continual Learning (CL) aims to build artificial intelligence (AI) systems capable of incrementally acquiring and integrating new knowledge while preserving previously learned information~\cite{van2022three, survey_wang}. This ability is fundamental for creating AI agents that can adapt to the dynamic and ever-changing real world, much like humans do~\cite{agi_for_cl_0, agi_for_cl_1}. However, contemporary AI, which is largely built upon deep neural networks, faces a significant obstacle known as catastrophic forgetting~\cite{mccloskey1989catastrophic,goodfellow2013empirical}. This phenomenon describes the tendency of a network's performance on prior tasks to degrade abruptly when it is trained on a new one. While simple strategies, such as restricting model updates, can mitigate forgetting, they often impair the model's ability to learn new things. This trade-off exposes the core stability-plasticity dilemma: the need to maintain stable knowledge representations while retaining the plasticity required to learn new information~\cite{grossberg2013adaptive}.

To navigate this dilemma, researchers have developed numerous CL methods. These approaches include regularizing changes to important parameters~\cite{ewc, chaudhry2018riemannian}, replaying data from past tasks~\cite{icarl, wangmemory, er}, and isolating dedicated parameters for different tasks~\cite{serra2018overcoming, yan2021dynamically}. In particular, certain works~\cite{van2020brain, gurbuz2022nispa} draw inspiration from natural continual learners, i.e., biological intelligence systems, to build the CL strategies~\cite{kudithipudi2022biological, hadsell2020embracing}. For example, Synaptic Intelligence (SI)~\cite{si} emulates the adaptive mechanisms of biological synapses to protect task-relevant weights, thereby mitigating forgetting. Another biologically inspired method~\cite{wang2023incorporating} actively manage forgetting by simulating memory attenuation and employing multi-learner architectures, mirroring processes observed in systems like the Drosophila brain.

Despite these advances, the prevailing CL paradigm remains focuses on the learning capacity of a single model. This approach, we argue, is inherently limited. Just as an individual human cannot master every skill, a single AI model has finite capacity for sequential learning a large number of tasks. Specifically, the optimal parameters or architectures for old and new tasks may conflict, learning all incremental tasks with a shared model inherently causes the inter-task interference~\cite{survey_wang}. Moreover, research has shown that models can gradually lose plasticity over successive learning phases, eventually impeding their capacity to acquire new knowledge~\cite{dohare2024loss,fariasself}.

Crucially, human intelligence does not emerge in isolation but thrives through the collective accumulation of knowledge across a population. In human societies, individuals specialize in different domains, contributing a diverse pool of expertise that drives the advancement of the entire group. Evolution further reinforces this dynamic by endowing individuals with complementary talents, making them naturally suited to different tasks. This evolutionary mechanism is what has enabled humanity to master increasingly complex skills across generations, from foundational discoveries like agriculture to modern technological innovations. It is this powerful principle of collective intelligence that motivates our work.

\begin{figure*}[!t]
    \centering
    \includegraphics[width=1\linewidth]{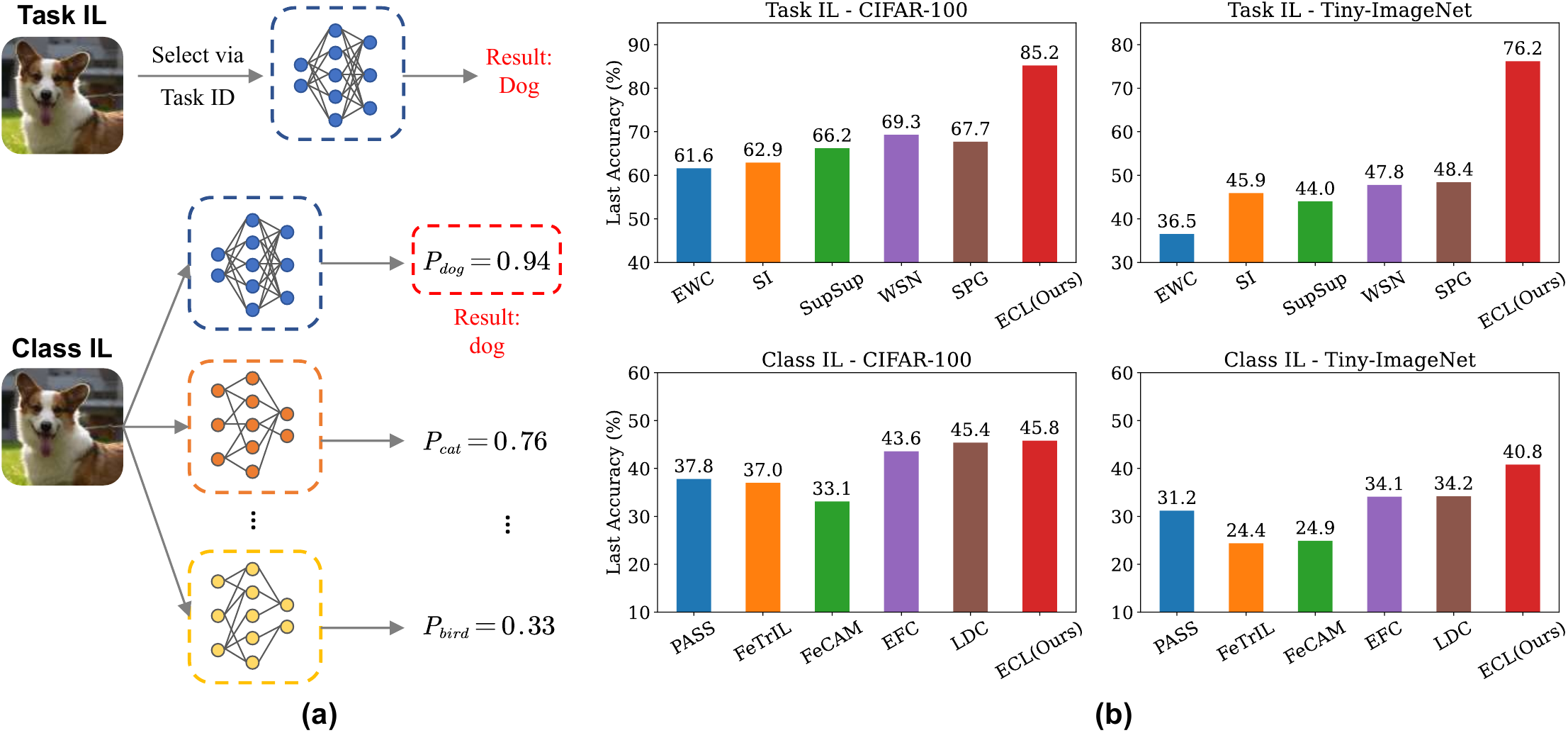}
    \caption{(a) Since the samples may come from any of the learned tasks during inference, we use two distinct inference strategies for Task IL and Class IL to select a task-specific model within ECL for decision. (b) In both settings, ECL achieves superior performance compared to existing individual-level CL methods across two typical datasets.}
    \label{fig:performance}
\end{figure*}

In this study, we propose Evolving Continual Learning (ECL), a paradigm that shifts the focus from single-model adaptation to population-based evolution. As shown in Figure~\ref{fig:fig1}, instead of relying on a single, unified model, ECL cultivates an evolving population of models to sequentially learn a variety of tasks. Each model in the population possesses a distinct architecture, making it potentially better suited for specific types of tasks. When a new task is introduced, the entire population of models evolves its architectures through an automated process of searching for neural network designs, known as Neural Architecture Search (NAS)~\cite{zoph2017neural, nas_survey}. The model demonstrating the highest potential during this architectural search process is selected to be trained on the new task. Once trained, this model is archived as a specialized "expert" for that specific task. The entire evolved population from the current stage is then passed on to serve as the starting point for the next task. This inheritance mechanism ensures that architectural refinements and adaptations are carried forward, inherently promoting forward knowledge transfer.

This iterative process yields a collection of specialized models, each with an architecture optimized for its designated task. Consequently, ECL elegantly addresses the stability-plasticity dilemma. Stability is intrinsically achieved through parameter isolation~\cite{yan2021dynamically, spg}, as different tasks are handled by distinct models. Simultaneously, plasticity is significantly enhanced by leveraging both tailor-made architectures and dedicated parameters for each new challenge. Our empirical evaluation demonstrates that ECL consistently and substantially outperforms existing individual-level CL methods across several standard benchmarks. To further validate our approach, we compared ECL to a baseline that uses an independent, non-specialized model for each task. The superior performance of ECL in this comparison confirms that the evolutionary search for optimized architectures is a critical component of its success.

\section{Results}

To validate the efficacy of ECL, we conduct a series of experiments designed to assess its performance against established baselines and to isolate the contributions of the core evolution process.

\begin{figure*}[!t]
    \centering
    \includegraphics[width=1\linewidth]{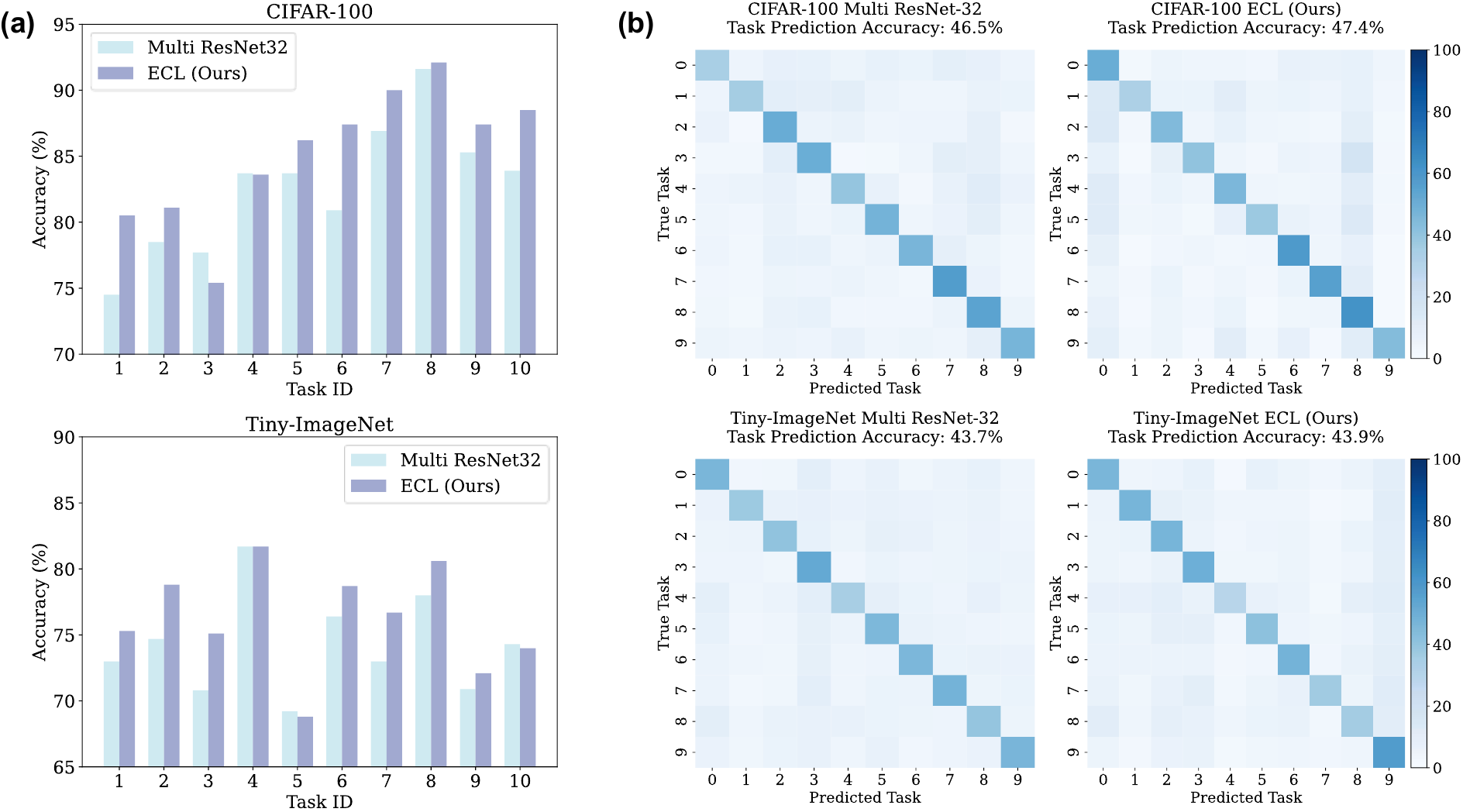}
    \caption{(a) Per-task accuracy comparison between ECL and a multi-ResNet-32 baseline, assuming known task identities. (b) Task-ID confusion matrices for ECL and the multi-ResNet-32 baseline. All results show that ECL outperforms baselines.}
    \label{fig:ablation}
\end{figure*}

\subsection{Comparison with Individual-level CL}
\label{subs:results}

A central challenge in CL is mitigating the mutual interference that occurs when a model learns a sequence of tasks with different data distributions~\cite{wang2023incorporating}. In human societies, this challenge is naturally addressed by assigning distinct responsibilities to individuals with corresponding talents. Drawing inspiration from this, we introduce ECL, formulating this process through an evolutionary framework. In ECL, a population of diverse network architectures, analogous to individuals with varied talents, continually evolves to generate a specialized model for each new task. The core objective of ECL is to leverage this evolutionary search to discover highly adapted and efficient network architectures for a sequence of tasks, thereby overcoming the limitations of conventional CL approaches.

ECL operates by evolving a population of candidate models over successive generations to optimize performance on the current task. This iterative process continues until a new task is introduced, at which point the population begins adapting to the new data distribution. This methodology results in a CL system composed of a collection of expert models, each tailored to a specific task. This paper evaluates ECL in two standard CL scenarios: Task-Incremental Learning (Task-IL), where task identities are provided during inference, and Class-Incremental Learning (Class-IL), where the model must infer the task identities during inference. The mechanism of ECL for selecting the corresponding model for inference is shown in Figure~\ref{fig:performance} (a). This process is straightforward in the Task-IL setting, i.e., the appropriate expert model is selected using the provided task identity. For the more challenging Class-IL scenario, where the task identity is unknown, the expert model that exhibits the highest prediction confidence for a given input is chosen to make the final decision.

Now, we evaluate the advantages of this population-based approach by comparing ECL with existing state-of-the-art CL techniques that rely on a single, unified model. The comparative performance results are presented in Figure~\ref{fig:performance} (b), which shows that ECL consistently and significantly outperforms all competing single-model methods across multiple datasets and both incremental settings. Specifically, in the Task-IL scenario, ECL achieves an improvement in Last Accuracy (LA) of 15.9\% on CIFAR-100 and 27.8\% on Tiny-ImageNet over the next-best method. The advantage holds in the Class-IL setting, where ECL surpasses the top competitor by 0.4\% on CIFAR-100 and 6.6\% on Tiny-ImageNet. These commanding results robustly validate the superiority of ECL's population-based evolutionary approach over the conventional single-model CL paradigm.

Notably, although ECL maintains a dedicated model for each task, it does not necessarily introduce greater memory overhead compared to existing CL approaches. This efficiency is achieved because each specialized model is designed to be highly parameter-efficient. Using a multi-objective evolutionary algorithm~\cite{deb2002fast}, ECL optimizes for both performance and model size. For instance, in our CIFAR-100 experiments, the cumulative size of all 10 final expert models is only 4.04M parameters. This is significantly smaller than the parameter counts of standard architectures commonly used in CL, such as ResNet-18 (11.2M), demonstrating the method's ability to achieve superior performance with parameter efficiency.

\subsection{Evolving Population vs. Multi Models}

After demonstrating the advantages of an evolving population over a single-model approach, we conduct a comparative analysis against a multi-model baseline to isolate the specific contributions of architectural evolution. This baseline consists of multiple independent ResNet-32 models, each trained separately on a single task while retaining a uniform architecture throughout the task sequence. This comparison separates the benefits of architectural specialization from those gained by using multiple models. Our analysis addresses two key research questions: (1) Does ECL’s architectural optimization improve per-task accuracy? (2) Does it enhance the reliability of task identity inference?

\textbf{Superior Per-Task Accuracy.} To evaluate performance on individual tasks, we first evaluate accuracy on each task with the correct task-specific model. As shown in Figure~\ref{fig:ablation} (a), the results demonstrate that ECL's evolved expert models consistently outperform the generic ResNet-32 baseline across most tasks on both datasets. These results confirm that the architecture optimization strategy, which tailors network designs to the specific distribution of each task's data, yields significant performance gains over fixed, generalized architectures.

\textbf{Task Discrimination.} A critical challenge in the Class-IL scenario is accurately inferring the task identity of a given input. Figure~\ref{fig:ablation} (b) presents the task confusion matrices for both methods, revealing that ECL exhibits lower inter-task misclassification rates across both datasets. This indicates that its specialized architectures are inherently more discriminative, making them more effective at inferring task identity from the input data alone. This improved discrimination is crucial for achieving high performance in realistic Class-IL settings, where task identity is not explicitly provided..

In summary, our analysis demonstrates that ECL’s task-specific architectural optimization delivers substantial benefits over approaches relying on fixed, generalized network designs. These findings highlight the importance of architectural evolution in CL, extending beyond the mere use of multiple models.

\section{Discussion}

In this study, we introduced ECL, a novel paradigm that shifts the focus of CL from individual model adaptation to population-based evolution. By leveraging an evolution-based NAS strategy, ECL automates the generation of highly specialized networks for each task in a sequence. Our empirical evaluation demonstrates that ECL delivers superior performance compared to state-of-the-art methods that rely on a single, unified model, establishing its potential as a powerful new framework.

The success of ECL stems from its innovative approach to resolving the stability-plasticity dilemma. It enhances plasticity by discovering tailored architectures for the current task while ensuring stability by archiving these models as independent experts, thereby preventing catastrophic forgetting. The evolutionary process itself serves as a mechanism for forward knowledge transfer, as the entire population's architectural adaptations are inherited by subsequent tasks. Moreover, a key advantage of this framework is that it is rehearsal-free, making it suitable for real-world scenarios where privacy concerns or logistical constraints prohibit access to past data. 

In conclusion, by emulating the principles of collective and specialized intelligence, this work opens a promising avenue for future research. We believe that exploring evolutionary and population-based dynamics is a key step toward building AI systems capable of sustained, robust, and truly open-ended learning.

\section{Method}

\subsection{Overall Framework of ECL}
\label{overall_framework}

The ECL process, formalized in Algorithm~\ref{alg:ecl}, begins with a randomly initialized population of architectures, $P$, drawn from a predefined search space. For each task $t$ in the sequence of $T$ tasks, the population undergoes an evolutionary optimization loop. This loop runs for a fixed number of generations, $G$, during which individuals in $P$ are trained for a few epochs and evaluated. Within each generation, parent architectures are selected based on their fitness, and new offspring are generated through crossover and mutation. The population for the next generation is then formed by selecting the highest-fitness individuals from the combined parent and offspring pools.

Upon the completion of the evolutionary search for task $t$, the best-performing architecture from the final population is identified. This optimal model, denoted as $Solution_t$, is then trained to full convergence on the training data for the current task, $D^t_{\text{train}}$, and subsequently archived. When a new task is introduced, the evolutionary process resumes, adapting the population to the new data distribution. The final output of the ECL process is a collection of expert models, each tailored to a specific task in the sequence.

\begin{algorithm}
\caption{Overall process of ECL}
\label{alg:ecl}
\begin{algorithmic}[1]
\Require Incremental datasets $D$ with $T$ tasks
\Ensure An optimal model population for CL tasks
\State $P \gets $ Initialize a population
\For{$t= 1, 2, \dots, T$}
    \State Train and evaluate the individuals in $P$
    \For{$g = 1, 2, \dots, G$}
        \State $P_{\text{parent}} \gets$ Select parents from $P$
        \State $Q \gets$ Generate new offspring via crossover and mutation
        \State Evaluate the individuals in $Q$
        \State Update $P$ by selecting high-fitness individuals from the previous $P$ and $Q$
    \EndFor
    \State $Solution^{}_t \gets$ Best-performing model in $P$
    \State Fine-tune $Solution^{}_t$ on $D^t_{\text{train}}$ and archive
\EndFor
\end{algorithmic}
\end{algorithm}

\begin{figure*}[ht]
    \centering
    \includegraphics[width=1.0\linewidth]{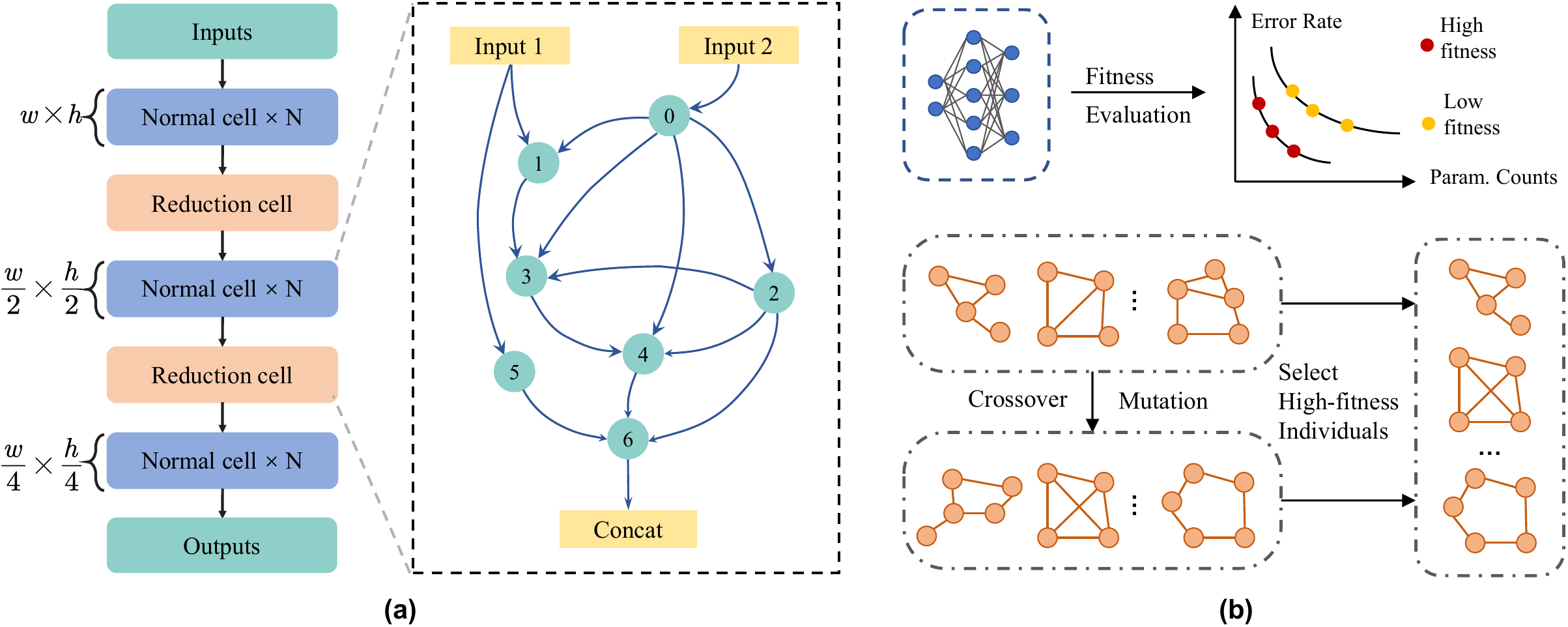}
    \caption{(a) Search Space of ECL. The entire architecture consists of several normal and reduction cells. A cell is a small network represented by a directed acyclic graph. (b) Search Strategy of ECL. The population generates offspring via crossover and mutation and the individual with high fitness (low error rate and low parameter counts) will be selected to the next generation. }
    \label{fig:methods}
\end{figure*}

\subsection{Architectural Search Space}
\label{subs:Search Space}

The design of the search space is critical to the efficacy of any NAS methods~\cite{radosavovic2020designing,wan2022redundancy}. The cell-based search space, renowned for its excellent scalability and efficiency, has gained considerable popularity~\cite{zoph2018learning,liudarts}. Drawing inspiration from this, we have crafted a search space for ECL that is also founded on the cell structure, with its overall design aligned with established methodologies like DARTS~\cite{liudarts}. 

As illustrated in Figure~\ref{fig:methods} (a), the network architecture is constructed by stacking learnable building blocks, termed "cells." These cells are categorized into normal and reduction cells, with the structure of cells of the same type defined as the same. Reduction cells are designed to halve the feature map's spatial dimensions, whereas normal cells preserve the original dimensions of the feature map. A cell is a directed acyclic graph in which each node represents an operation (e.g., convolution). And if there is a connection between nodes $i$ and $j$ ($i < j$), it means that the output of node $i$ is fed into node $j$. The NAS process is primarily concerned with the seamless exploration of architectural configurations within these cells, searching for the optimal operation type and connections for each node.

Specifically, our search space encompasses seven distinct operations: $3 \times 3$ and $5 \times 5$ separable convolutions, $3 \times 3$ and $5 \times 5$ dilated convolutions, $3 \times 3$ max pooling, $3 \times 3$ average pooling, and an identity operation. Furthermore, recognizing that optimal network depth is task-dependent and cannot be predetermined in a CL scenarios~\cite{mirzadeh2022wide, CL_design}, we allow the number of nodes within each cell to be a variable in the search. Specifically, the number of nodes in both normal and reduction cells can range from 5 to 12, enabling the evolutionary algorithm to automatically discover a suitable network depth for each sequential task.

\subsection{Search Strategy and Performance Evaluation}

For the search strategy, ECL employs a multi-objective evolutionary algorithm inspired by NSGA-II~\cite{deb2002fast}. This approach is designed to balance the often-competing goals of performance and efficiency, as depicted in Figure~\ref{fig:methods} (b). Each candidate architecture is evaluated against two fitness objectives: (i) Validation Error Rate: The primary performance metric, computed as the classification error rate on the validation set of the current task. (ii) Model Complexity: The total number of parameters in the model, serving as a proxy for efficiency and resource constraints. The final fitness is mainly decided by their Pareto rank assigned via non-dominated sorting. Within each front, individuals with higher crowding distances (i.e., less densely populated regions of the objective space) are preferred to promote diversity.

For each incoming task, a population of candidate architectures is randomly initialized or inherited from the previous task. The evolutionary process proceeds in iterative cycles. Firstly, parent individuals are selected from the current population using a binary tournament based on their fitness. Next, crossover and mutation operators are applied to the selected parents to generate an offspring population. Finally, the parent and offspring populations are combined, and environmental selection is performed to select the most fit individuals for the next generation's population.

To get the validation error rate, a full training cycle for every candidate architecture would be effective but computationally prohibitive. We therefore adopt an efficiency evaluation strategy based on the observation that early-stage performance is highly correlated with final converged performance~\cite{PEPNAS}. During the search phase, candidate architectures are trained for only a small number of epochs to obtain a computationally cheap estimate of their validation error rate. Once the search for a given task is complete, the single top-performing architecture from the final Pareto front (i.e., the one with the lowest validation error) is selected and trained to full convergence to serve as the specialized model for that task.

\subsection{Experimental Setting}
\label{subs:setting}

\textbf{Baseline Approaches.} We evaluate ECL against state-of-the-art individual-level CL approaches under the replay-free setting. For Task IL evaluation, we select five representative methods as baselines: EWC~\cite{ewc}, SI~\cite{si}, SupSup~\cite{supsup}, WSN~\cite{wsn}, and SPG~\cite{spg}. For Class IL evaluation, we compare ECL with PASS~\cite{pass}, FeTrIL~\cite{fetril}, FeCAM~\cite{fecam}, EFC~\cite{efc}, and LDC~\cite{ldc}. Following established conventions~\cite{spg, ldc}, all Task IL baselines use AlexNet as the backbone architecture, while ResNet-18 is employed for Class IL baselines. The baseline results are sourced from prior studies~\cite{spg, ldc}.

\textbf{Datasets.} In line with prior work~\cite{spg, ldc}, we conduct experiments on two widely-used benchmark datasets: CIFAR-100~\cite{cifar100} and Tiny-ImageNet~\cite{le2015tiny}. Each dataset is divided into 10 sequential tasks. For CIFAR-100, each task consists of 10 non-overlapping classes, while for Tiny-ImageNet, each task contains 20 non-overlapping classes. This setup ensures a consistent and challenging evaluation environment for both Task IL and Class IL scenarios.

\textbf{Implementation Details.} To ensure efficient architecture search, we adopt a lightweight configuration with an initial channel width of 16 (doubled after each reduction cell) and a shallow network structure defined by $N=1$, resulting in architectures comprising 3 normal cells and 2 reduction cells. For the final evaluation, we enhance model capacity by increasing the channel width to a fixed size of 64 and deepening the network with $N=3$, yielding 9 normal cells and 2 reduction cells.

The evolutionary search employs a population size of 20 and runs for a maximum of 10 generations in the first task, followed by 5 generations for each subsequent task. All models are optimized using Stochastic Gradient Descent (SGD) with a momentum of 0.9 and a weight decay coefficient of 0.0005. The learning rate follows a cosine annealing schedule, initialized at 0.1. During the fitness evaluation phase, models undergo training for 10 epochs, with the learning rate bounded by a lower limit of 0.001. Once the optimal architecture is selected for each task, we conduct full training for 200 epochs to maximize performance. In this phase, the learning rate is further constrained to a minimum of 1e-5 to ensure fine-grained convergence.

\textbf{Evaluation Metrics.} The primary metric for assessing CL performance in both Task-IL and Class-IL scenarios is Last Accuracy (LA). This metric captures the model's ability to retain knowledge from past tasks while learning new ones. It is calculated as the average classification accuracy across all encountered tasks after the model has been trained on the final task. Formally, let $K$ denote the total number of tasks, and let $a_{i}$ represent the classification accuracy evaluated on the test set of the $i$-th task after completing all $K$ tasks. The LA is defined as:
\[
LA=\frac{1}{K} \sum_{i=1}^{K} a_{i}
\]
Higher LA values signify superior CL performance, i.e., a good balance between stability and plasticity.

\bibliography{sn-bibliography}

\end{document}